\documentclass[11pt]{article}
\usepackage[]{acl}

\usepackage{times}
\usepackage{latexsym}
\usepackage[T1]{fontenc}
\usepackage[utf8]{inputenc}
\usepackage{microtype}
\usepackage{inconsolata}

\usepackage{amsmath, amssymb}
\usepackage{mathtools}

\usepackage{graphicx}
\usepackage{booktabs}
\usepackage{multirow}
\usepackage{tabulary}
\usepackage{diagbox}
\usepackage{color, colortbl}
\usepackage{caption}
\newcommand{\rev}[1]{\textcolor{black}{#1}}

\usepackage{tikz}
\usepackage{pgfplots}
\pgfplotsset{compat=newest}

\usepackage{enumitem}
\usepackage{tcolorbox}
\tcbuselibrary{breakable}
\usepackage{listings}

\usepackage[nolist,nohyperlinks]{acronym}

\definecolor{Gray}{gray}{0.95}
\definecolor{LightCyan}{rgb}{0.8,0.85,1}
\definecolor{viridisStart}{rgb}{0.3137, 0.1412, 0.4549}
\definecolor{viridisEnd}{rgb}{0.1882, 0.4549, 0.5490}

\setlist{nosep}

\title{TeleTables: A Benchmark for Evaluating LLM Understanding\\
       of Tables in Technical Standards}

\author{Anas Ezzakri$^{1}$\quad Nicola Piovesan$^{1}$\quad Mohamed Sana$^{1}$\quad
Antonio De Domenico$^{1} $\quad \\\textbf{Fadhel Ayed}$^{1}$\quad \textbf{Haozhe Zhang}$^{2}$ \\[2pt]
$^{1}$Paris Research Center, Huawei Technologies, Boulogne-Billancourt, France \\
$^{2}$Shanghai Research Center, Huawei Technologies, Shanghai, China
}

\begin{document}
\maketitle

%%%%%%%%%%%%%%%%%%%%%%%%%%%%%%%%%%%%%%%%%%%%%%%%%%%%%
\begin{abstract}
\rev{Large Language Models (LLMs) are increasingly applied to telecom engineering tasks, yet perform poorly on 3GPP specifications. These standards encode much of their technical information in complex tables, but LLM knowledge and interpretation of such tables remain largely unexplored.}

\rev{We introduce \textbf{TeleTables}, a benchmark comprising 2,220 tables from 13 3GPP specifications in four formats and 500 human-verified MCQs spanning direct retrieval to multi-step reasoning.}

\rev{Evaluating 20 open-weight LLMs across non-reasoning, multimodal, reasoning, and table-specialized architectures reveals two distinct performance bottlenecks. In the closed-book setting, domain knowledge is the primary constraint, with no general-purpose model exceeding 41\% accuracy. When the table is provided as context, the best models exceed 90\%, but performance degrades systematically with reasoning depth, evidence scope, and structural complexity, with a 32.2\,pp spread across reasoning skills.}

\rev{Table specialization on non-telecom data provides no consistent benefit, while strong reasoning capabilities remain essential for reliable interpretation of complex technical tables.}

\end{abstract}
%%%%%%%%%%%%%%%%%%%%%%%%%%%%%%%%%%%%%%%%%%%%%%%%%%%%%
\section{Introduction}
\label{sec:intro}
\textcolor{black}{Structured tables are a primary vehicle for encoding information in technical standards documents. In domains such as telecommunications, finance, avionics, medicine, specifications routinely contain hundreds of tables per document, with multi-level headers, merged cells, cross-table references, and condition-dependent entries that encode relationships not expressible in plain text.
Reliable table understanding is therefore a prerequisite for \ac{LLM}-driven automation of practical engineering tasks: automated configuration, parameter look-up, and specification-driven troubleshooting all depend on correctly reading these structures.
Yet despite the ubiquity and complexity of such tables, they remain largely absent from existing NLP benchmarks for table question answering.}

\textcolor{black}{The telecommunications domain, where \ac{3GPP} specifications form the backbone of network engineering, offers a concrete and high-stakes testbed for this challenge.}
\acp{LLM} have the potential to assist in a wide range of telecom tasks, from troubleshooting network anomalies~\citep{sana2025reasoning} to interpreting standards and recommending configuration strategies~\citep{llm:telecom_impact}.
\textcolor{black}{Yet existing evaluations already reveal that 3GPP-related questions are the hardest class for LLMs:} TeleQnA~\citep{teleqna:paper}\textcolor{black}{, a large-scale telecom benchmark}, found that LLMs \textcolor{black}{score lowest on standards-based questions}.
\ac{RAG} systems tailored to standard documents~\citep{chat3gpp} improve accuracy by up to 18\%, yet 3GPP questions remain the weakest category overall~\citep{telcoorag2025}.
\textcolor{black}{We argue that a key unexamined cause is the heavy reliance on tables: a typical 3GPP specification contains several hundred tables with nested headers, merged cells, and hierarchical structures that existing LLMs and benchmarks are ill-equipped to handle (Appendix \ref{app:tabex}). Existing table QA benchmarks focus on general-domain data such as Wikipedia or business intelligence tables, and none examines whether their findings transfer to technical standards.}

\paragraph{Contributions.}
\begin{itemize}[leftmargin=*]
    \item \textcolor{black}{\textbf{Table collection.} We publicly release a corpus of 2,220 tables from 13 3GPP specifications in four formats (HTML, JSON, Markdown, image). This is the first multi-format collection of tables from a technical standards domain.}
    \item \textcolor{black}{\textbf{Benchmark.} We curate} TeleTables\footnote{\url{https://huggingface.co/datasets/netop/TeleTables}}, \textcolor{black}{500 human-verified \acp{MCQ} paired with their corresponding 3GPP tables, spanning simple retrieval to multi-step reasoning, and designed to evaluate both implicit LLM knowledge and explicit table interpretation ability.}
    \item \textcolor{black}{\textbf{Pipeline.} We design a two-stage, LLM-based MCQ generation framework that combines multimodal generation, multi-trial validation, a difficulty filter, and human review. The framework is reusable for benchmark construction from any technical document corpus.}
    \item \textcolor{black}{\textbf{Findings.} We evaluate 20 open-weight LLMs across model types and table formats.} Our evaluation shows that HTML is the strongest single representation for most models and that reasoning models consistently outperform table-specialized models, regardless of size. \rev{We further identify two distinct performance bottlenecks: in the closed-book setting, domain knowledge is the primary constraint, with no general-purpose model exceeding 40.6\% cons@16; when the table is provided as context, general-purpose models reach 85--92\%, but table interpretation becomes the primary constraint, with accuracy systematically decreasing with reasoning depth, evidence scope, and structural complexity.}
\end{itemize}

%%%%%%%%%%%%%%%%%%%%%%%%%%%%%%%%%%%%%%%%%%%%%%%%%%%%%
\section{Related Work}
\label{sec:related}

\textcolor{black}{\paragraph{Table question answering.}
\citet{sui2024table} provided one of the first systematic analyses of LLM table understanding; TableBench~\citep{wu2025tablebench} and TableEval~\citep{zhu2025tableeval} extended this to multi-step reasoning and multilingual settings, finding consistent struggles with hierarchical formats. \citet{zheng2024multimodal} benchmark vision-language models on image-rendered tables, finding additional failure modes beyond text-based parsing. None of these targets tables from technical standards, which combine domain-specific terminology, deeply nested structure, and cross-document referencing.}

\textcolor{black}{\paragraph{Table-specialized models.}
TableGPT2~\citep{su2024tablegpt2} applies continual pretraining and supervised fine-tuning on large collections of structured data, achieving strong results on business-intelligence table tasks.
More recent approaches apply reinforcement learning: Table-R1~\citep{yang2025tabler1} uses verifiable reward signals derived from table lookups, while TableGPT-R1 extends this paradigm.
Our evaluation shows that specialization on these general-domain corpora does not transfer to 3GPP tables, and in some cases degrades performance relative to the unmodified base model.}

\textcolor{black}{\paragraph{LLM benchmarks for telecommunications.}
TeleQnA~\citep{teleqna:paper} introduced the first large-scale telecom MCQ benchmark, revealing that 3GPP standards-related questions are the hardest category for all evaluated models.
TSpec-LLM~\citep{nikbakht2024tspec} released an open-source dataset for LLM understanding of 3GPP specifications, while TeleMath~\citep{colle2024telemath} targets mathematical problem solving in the telecom domain.
Closest to our work, \citet{roychowdhury2024evaluation} evaluated different table representations for answering questions from 3GPP documents, finding that representation format matters. TeleTables differs by providing a larger, human-verified benchmark with a systematic difficulty stratification, a multi-format table collection released as a standalone resource, and an evaluation of both closed-book knowledge and in-context table understanding across a broader set of model types including table-specialized architectures.}

%%%%%%%%%%%%%%%%%%%%%%%%%%%%%%%%%%%%%%%%%%%%%%%%%%%%%
\section{\textcolor{black}{Table Collection}}
\label{sec:sources}
\textcolor{black}{We construct a corpus of 2,220 tables from 13 3GPP technical specifications (Release 18--19, spanning the TS 23.xx, 36.xx, and 38.xx series), released in four synchronized formats: HTML, JSON, Markdown, and PNG image.
Tables were extracted via a four-stage pipeline: (i)~\textit{localization}, parsing each document for elements matching the standard table title convention; (ii)~\textit{extraction} using Pandoc to produce HTML, from which JSON and Markdown are derived, with a parallel high-resolution PNG preserving the visual layout including merged cells and hierarchical headers; (iii)~\textit{metadata collection}, enriching each table with its identifier, caption, and relevant contextual paragraphs; and (iv)~\textit{LLM-based refinement} using Qwen2.5-VL-72B-Instruct to transcribe non-textual elements, verify format fidelity against the image, and filter irrelevant context.}

\begin{figure*}[t]
    \centering
    \includegraphics[width=.9\textwidth, trim={0.8cm 0 0cm 0.2cm},clip]{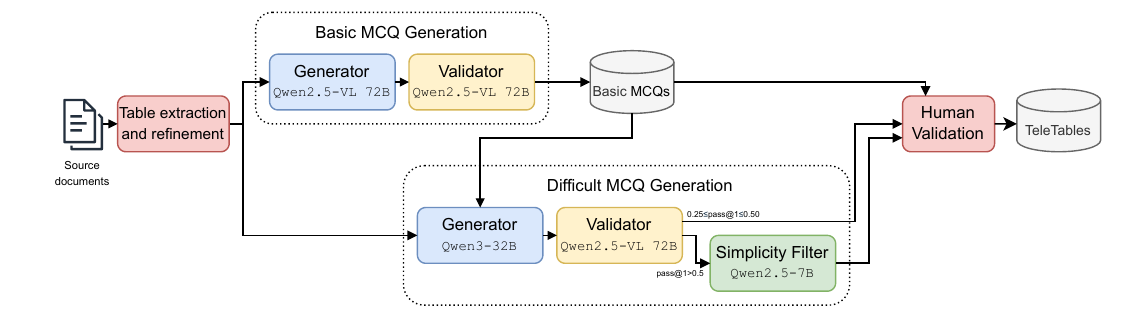}
    \caption{Overview of the MCQ generation pipeline.}
    \label{fig:difficult_subset_pipeline}
\end{figure*}

%%%%%%%%%%%%%%%%%%%%%%%%%%%%%%%%%%%%%%%%%%%%%%%%%%%%%
\section{Dataset Characteristics}

\rev{\subsection{Table Corpus}
Table~\ref{tab:corpus-taxonomy} reports the distribution of structural and content properties across the annotated corpus.
The corpus exhibits substantial structural complexity: 28.4\% of tables feature multi-level headers, 13.1\% contain grouped rows with merged cells, and 38.3\% include footnotes that qualify or constrain cell values.
Table content is also heterogeneous: 76.2\% of tables contain numeric or parametric data, 43.0\% include prose descriptions, and 30.8\% contain mathematical expressions.
These characteristics reflect the heterogeneous structure and specialized content of tables found in technical standards, and distinguish the corpus from general-domain table benchmarks, which typically comprise clean single-type tables.}

\begin{table}[t]
\centering
\small
\caption{\rev{Distribution of structural and content properties across the TeleTables corpus.}}
\label{tab:corpus-taxonomy}
\resizebox{\columnwidth}{!}{%
\begin{tabular}{llr}
\toprule
\textbf{Dimension} & \textbf{Category} & \textbf{\%} \\
\midrule
\multirow{3}{*}{Table size}
  & Small ($\leq$5 rows, $\leq$4 cols) & 30.4\% \\
  & Medium (6--20 rows or 5--8 cols)   & 48.4\% \\
  & Large ({>}20 rows or {>}8 cols)   & 21.2\% \\
\addlinespace
\multirow{2}{*}{Header structure}
  & Flat (single row)                        & 71.6\% \\
  & Multi-level (colspan / rowspan)         & 28.4\% \\
\addlinespace
\multirow{2}{*}{Row structure}
  & Flat rows (no merging)                  & 86.1\% \\
  & Grouped rows (rowspan)               & 13.1\% \\
\addlinespace
\multirow{5}{*}{Cell content$^{\dagger}$}
  & Numeric / parametric                    & 76.2\% \\
  & Prose descriptions                        & 43.0\% \\
  & Categorical labels / codes              & 34.2\% \\
  & Mathematical expressions                 & 30.8\% \\
  & Symbol / bit sequences                    & 2.7\%  \\
\addlinespace
\multirow{2}{*}{Footnotes}
  & With notes / qualifiers                  & 38.3\% \\
  & Without notes                           & 61.7\% \\
\bottomrule
\end{tabular}}\\[2pt]
{\footnotesize $^\dagger$ Multi-label: tables may belong to multiple content-type categories.}
\end{table}

\rev{\subsection{Questions}
TeleTables adopts the \ac{MCQ} format as it enables objective, automated evaluation and allows nuanced distractors that probe a model's ability to discriminate between subtly different interpretations of tabular content \cite{mctest2013}.
The benchmark comprises 500 human-verified \acp{MCQ}, of which 250 are classified as \textit{Difficult} (requiring multi-step or aggregated reasoning).
Each question is annotated along two dimensions: \textit{reasoning skill} (Q-Lookup, Q-Count, Q-Arithmetic, Q-Compare, Q-Average, Q-MultiCriteria) and \textit{evidence scope} (E-Cell, E-Cross, E-Full), which underpin the stratified analysis in Section~\ref{sec:stratified}.}

%%%%%%%%%%%%%%%%%%%%%%%%%%%%%%%%%%%%%%%%%%%%%%%%%%%%%
\section{Dataset Creation}
\label{sec:dataset_creation}

We developed a two-stage pipeline for MCQ generation followed by human validation, illustrated in Figure~\ref{fig:difficult_subset_pipeline}.

\subsection{Basic MCQ Generation}
\label{subsec:foundational_generation}
The first stage of the pipeline is designed to produce a large volume of basic \acp{MCQ} covering the breadth of the extracted tables. 

A generator agent (Qwen2.5-VL-72B-Instruct) is provided with all four table representations and tasked with producing \acp{MCQ} based on the table content. To mitigate the LLM positional bias, the answer choices are randomly shuffled once before validation.

The generated MCQs are then independently validated by a second agent.
The validator receives the MCQ and the corresponding table as input, and it is tasked 
with independently selecting the correct option.
To avoid any bias, the correct answer is omitted from the input. 
Validation is performed across four independent trials, and an \ac{MCQ} is retained only if the correct option is selected in at least three trials; otherwise, it is discarded as ambiguous, inconsistent, or erroneous.

\subsection{Difficult MCQ Generation}
The second stage of the pipeline aims to increase the MCQ complexity by transforming the validated basic questions into more advanced reasoning-oriented ones.
For each table, the generator is provided with five randomly selected basic \acp{MCQ} together with all table formats.
It then synthesizes concepts from these simpler questions to construct more complex MCQs that require multi-step reasoning and the aggregation of intermediate results to derive the final answer.
This approach is inspired by structured reasoning paradigms such as Chain-of-Thought prompting~\citep{wei2022chain}. To support this capability, the generator agent is implemented using the Qwen3-32B reasoning model.

Difficult MCQs are validated by Qwen2.5-72B-Instruct across four trials. The number of successful validation trials serves as a quality indicator:
\begin{itemize}
    \item \textbf{0 successful trials:} the \ac{MCQ} is deemed fundamentally flawed and discarded.
    \item \textbf{1–2 successful trials:} the \ac{MCQ} may either contain errors or be excessively difficult to answer consistently. Such cases are flagged for human validation.
    \item \textbf{3–4 successful trials:} the \ac{MCQ} is considered correct and retained.
\end{itemize}
To ensure the difficult set of MCQs does not include trivially easy questions, a simplicity filter is applied. This filter is implemented using a smaller model (i.e., Qwen2.5-7B-Instruct), which attempts to answer each question across four trials. If the model consistently selects the correct answer in the majority of trials, the \ac{MCQ} is labeled as too easy and discarded; otherwise, it is preserved and forwarded to the human validation stage.

\subsection{Human-in-the-Loop Validation}
The final stage incorporates human feedback to guarantee the reliability and accuracy of the generated \acp{MCQ}.
Candidate MCQs are drawn from three sources: (i) basic questions produced in the first stage, (ii) questions identified as difficult by the simplicity filter in the second stage, and (iii) questions flagged by the validator as requiring manual inspection in the second stage.

\rev{Review was performed by 8 expert annotators with direct experience of 3GPP standards. An MCQ was accepted only if it was (i) factually correct given the table content, (ii) representative of a question that a domain expert would plausibly formulate, and (iii) unambiguous, i.e., exactly one option was defensible. Only questions satisfying all three criteria were retained.}

\rev{The full generation funnel was: 3,000 MCQs generated $\to$ 1,412 passed validator filtering $\to$ 817 passed human review $\to$ 500 sampled for public release.}
\rev{Examples of accepted and rejected questions are provided in Appendix~\ref{app:rejected}.}

%%%%%%%%%%%%%%%%%%%%%%%%%%%%%%%%%%%%%%%%%%%%%%%%%%%%%
\section{Can LLMs Answer Questions About 3GPP Tables?}
In this section, we first evaluate the ability of open-weight \acp{LLM} to answer TeleTables \acp{MCQ} without access to the associated tables or to the internet, relying solely on a textual reference to the relevant source. In this setup, a model can answer a question correctly only if it was exposed to the corresponding standard document during pretraining and has learned an effective internal representation of the table. The prompt explicitly refers to the table by instructing the model to answer the question \textit{``considering the table \textless table\_id\textgreater\ from 3GPP document \textless document\_title\textgreater.''}

Each model generates $N=16$ independent responses per MCQ using a sampling temperature of 0.6, a top-$p$ value of 0.90, and a maximum generation length of 16,384 tokens.
We assess the model performance using two complementary metrics: \textit{pass@1} and \textit{cons@16}.
The \textit{pass@1} metric measures the ability of the model to generate the correct answer in a single attempt, averaged over multiple samples, while \textit{cons@16} evaluates consistency through majority voting across 16 generated answers, considering an MCQ correctly answered if the most frequent response is correct.

\begin{table}[t]
\centering
\small
\caption{Performance of open-weight models on TeleTables (no table provided).}
\label{tab:knowledge}
\resizebox{\columnwidth}{!}{%
\begin{tabular}{llcc}
\toprule
\textbf{Model Type} & \textbf{Model} & \textbf{pass@1} & \textbf{cons@16} \\
\midrule
\multirow{5}{*}{\textbf{Non-reasoning}}
& Qwen2.5-7B-Instruct            & 26.15 & 26.00 \\
& Llama3.1-8B-Instruct           & 19.53 & 17.00 \\
& Qwen2.5-32B-Instruct           & 32.46 & 35.00 \\
& Llama3.3-70B-Instruct          & 33.24 & 34.80 \\
& Qwen2.5-72B-Instruct           & 31.08 & 32.20 \\
\midrule
\multirow{3}{*}{\textbf{Multimodal}}
& Gemma-27B-it                   & 32.25 & 32.20 \\
& Qwen2.5-VL-32B-Instruct        & 29.71 & 30.00 \\
& Qwen2.5-VL-72B-Instruct        & 32.72 & 36.20 \\
\midrule
\multirow{6}{*}{\textbf{Reasoning}}
& Qwen3-4B                       & 31.67 & 32.00 \\
& GPT-OSS-20B                    & 32.76 & 37.20 \\
& Qwen3-32B                      & 37.53 & 39.40 \\
& SeedOSS-36B-Instruct           & 34.56 & 37.40 \\
& DeepSeek-R1-Distill-Llama-70B  & 34.54 & 35.40 \\
& GPT-OSS-120B                   & 38.19 & 40.60 \\
\midrule
\multirow{1}{*}{\textbf{Domain fine-tuned}}
& \textcolor{black}{Qwen2.5-7B (SFT)}  & \textcolor{black}{\textbf{50.48}} & \textcolor{black}{\textbf{58.00}} \\
\bottomrule
\end{tabular}
}
\end{table}

Table~\ref{tab:knowledge} summarizes the performance of non-reasoning, multimodal, and reasoning \acp{LLM} when evaluating their knowledge of 3GPP tables.
Models achieve an average of roughly 35\% cons@16, with the lowest performance for Llama3.1-8B-Instruct (17\%) and the highest for GPT-OSS 120B (40.6\%).

These low accuracies indicate that open-weight \acp{LLM} are largely unable to answer questions related to 3GPP standard tables, suggesting that such material is either underrepresented in their pretraining data or the LLMs fail to correctly interpret it.

\subsection{Effect of Domain-Specialized Fine-Tuning}
\label{sec:sft}
The low zero-context scores raise a natural question: can targeted training on 3GPP material close the gap?
To investigate this, we fine-tuned Qwen2.5-7B-Instruct on a broad collection of 3GPP tables and documents paired with diverse task types, including table QA, table completion, structure extraction, and multi-step reasoning. \rev{The SFT corpus includes tables that also appear in TeleTables, but only for non-QA tasks; no TeleTables benchmark questions or answers are included in the training data.}

Fine-tuning yields substantial gains: pass@1 rises from 26.15\% to 50.48\% and cons@16 from 26.00\% to 58.00\% (Table~\ref{tab:knowledge}).

The fine-tuned 7B model outperforms all tested open-weight models in the zero-context evaluation, including GPT-OSS-120B (40.60\% cons@16), despite being approximately 17$\times$ smaller. This result indicates that, in the zero-context setting, domain-specific training can outweigh model scale.

%%%%%%%%%%%%%%%%%%%%%%%%%%%%%%%%%%%%%%%%%%%%%%%%%%%%%
\section{Do LLMs Understand 3GPP Tables?}

Building on the previous results, we now examine whether the limitation lies in access to the information or in the ability of the models to interpret it.
We evaluate LLMs with the relevant 3GPP table provided alongside each MCQ, mimicking a \ac{RAG} setup with perfect retrieval \textcolor{black}{(i.e., an upper bound on retrieval-augmented performance)}, isolating table understanding from pretraining exposure.

\subsection{Impact of Table Representation Formats}
Full per-format results are reported in Table~\ref{tab:input_formats} (Appendix~\ref{app:formats}); here we summarise the key findings.
HTML yields the strongest performance for most models which we attribute to its explicit markup of hierarchical cell relationships.
Llama models are an exception, performing marginally better with Markdown, likely reflecting their pretraining distribution.
Multi-format concatenation provides modest gains for small models (+5.2pp for Llama3.1-8B with JSON+MD) but negligible improvement for larger ones.
Image representations perform substantially worse (Qwen2.5-VL-72B: 77.2\% vs.\ 85.6\% with HTML), and hybrid image+text inputs recover most of this loss at higher token cost (12,831 tokens).
Markdown offers the best efficiency trade-off, using 42\% fewer tokens than HTML at competitive accuracy.

\subsection{Performance Analysis of Open-Weight Models}
In this section, we present a detailed analysis of the performance of non-reasoning, multimodal, reasoning, and table-specialized LLMs on the TeleTables benchmark.
Table~\ref{tab:model_types} reports how model scale, architecture, and reasoning capability impact accuracy.
For each model, results correspond to the best-performing input format or combination.
Accuracy is reported on both the full dataset and the subset of Difficult questions.

\begin{table*}[t]
\centering
\small
\caption{Performance of different models by type and best table representation format. Results are reported for the overall MCQs and for the \textit{Difficult} subset using \textit{pass@1} and \textit{cons@16}.}
\label{tab:model_types}
\resizebox{0.9\textwidth}{!}{%
\begin{tabular}{llccccc}
\toprule
\textbf{Model Type} & \textbf{Model} & \textbf{Best Format} &
\multicolumn{2}{c}{\textbf{Overall}} &
\multicolumn{2}{c}{\textbf{Difficult}} \\
\cmidrule(lr){4-5} \cmidrule(lr){6-7}
& & & \textbf{pass@1} & \textbf{cons@16} & \textbf{pass@1} & \textbf{cons@16} \\
\midrule
\multirow{5}{*}{\textbf{Non-reasoning}}
& Qwen2.5-7B-Instruct            & JSON         & 48.08 & 49.40 & 31.43 & 33.20 \\
& Llama3.1-8B-Instruct           & JSON+MD      & 56.50 & 62.60 & 36.00 & 44.00 \\
& Qwen2.5-32B-Instruct           & HTML         & 82.53 & 85.60 & 66.73 & 72.00 \\
& Llama3.3-70B-Instruct          & HTML+JSON    & 80.16 & 82.40 & 64.40 & 68.80 \\
& Qwen2.5-72B-Instruct           & HTML         & 76.66 & 79.40 & 65.85 & 71.20 \\
\midrule
\multirow{3}{*}{\textbf{Multimodal}}
& Gemma-27B-it                   & HTML+MD      & 80.45 & 82.20 & 64.08 & 67.20 \\
& Qwen2.5-VL-32B-Instruct        & IMAGE+HTML   & 83.04 & 86.80 & 67.38 & 74.40 \\
& Qwen2.5-VL-72B-Instruct        & HTML+JSON    & 84.46 & 87.40 & 71.45 & 76.80 \\
\midrule
\multirow{7}{*}{\textbf{Reasoning}}
& Qwen3-4B                       & HTML+JSON+MD & 86.26 & 88.60 & 76.50 & 80.40 \\
& Qwen3-8B                       & HTML+JSON+MD & 88.48 & 89.60 & 79.80 & 82.00 \\
& GPT-OSS-20B                    & HTML+JSON+MD & 88.04 & 90.20 & 78.65 & 82.40 \\
& Qwen3-32B                      & HTML+JSON+MD & \textbf{91.18} & \textbf{92.60} & \textbf{83.60} & \textbf{86.40} \\
& SeedOSS-36B-Instruct           & HTML+JSON    & 90.91 & 91.60 & 82.98 & 84.00 \\
& DeepSeek-R1-Distill-Llama-70B  & HTML+MD      & 88.86 & 90.60 & 80.10 & 83.20 \\
& GPT-OSS-120B                   & HTML+JSON+MD & 90.90 & 91.80 & 83.40 & 85.20 \\
\midrule
\multirow{5}{*}{\textbf{Table-Specialized}}
& TableGPT2-7B                   & Plain-text   & 41.83 & 46.40 & 28.13 & 34.00 \\
& Table-R1-Zero-7B               & HTML         & 69.25 & 73.00 & 47.23 & 53.20 \\
& Table-R1-Zero-8B               & HTML         & 61.66 & 64.80 & 42.35 & 45.20 \\
& TableGPT-R1                    & Plain-text   & 83.70 & 87.20 & 72.13 & 77.60 \\
& STaR-8B                        & MD           & 84.40 & 88.20 & 73.25 & 80.00 \\
\midrule
\multirow{1}{*}{\textbf{Domain fine-tuned}}
& \textcolor{black}{Qwen2.5-7B (SFT)}  & HTML & \textcolor{black}{59.00} & \textcolor{black}{65.00} & \textcolor{black}{36.38} & \textcolor{black}{45.20} \\
\bottomrule
\end{tabular}
}
\end{table*}

\subsubsection{Non-Reasoning Models}
Among non-reasoning models, performance increases with model size.
Smaller models (below 10B parameters) fail to exceed 75\% in cons@16, lacking the capacity to interpret and reason over technical specification tables.
Larger models surpass 79\% cons@16 overall, with Qwen2.5-32B 
reaching 85.60\% cons@16, though accuracy drops to 72.00\% on the 
Difficult subset, indicating that even large non-reasoning architectures 
struggle with multi-step tabular reasoning and arithmetic aggregation.

\subsubsection{Multimodal Models}
Multimodal LLMs exhibit superior performance compared to non-visual counterparts, demonstrating a robust capacity to interpret structured tabular data, benefiting from cross-modal pretraining even in the absence of image inputs.
For example, the best multimodal model, Qwen2.5-VL-72B, achieves 84.46\% pass@1 and 87.40\% cons@16 when using HTML and JSON inputs, outperforming its text-only counterpart Qwen2.5-72B (76.66\% pass@1, 79.40\% cons@16). This improvement is due to the enhanced representational capacity of multimodal models, which leverage vision-language alignment mechanisms that generalize beyond purely visual understanding.
Moreover, the extended context windows of multimodal architectures allow for the integration of multiple complementary table formats (e.g., HTML+JSON or Image+HTML), further enhancing their accuracy.

\subsubsection{Reasoning Models}
Reasoning models consistently achieve the highest performance on TeleTables, demonstrating their superior capacity to extract, aggregate, and reason over structured tabular information. The best-performing model, Qwen3-32B, reaches 91.18\% pass@1 and 92.60\% cons@16 overall, and maintains strong performance on the Difficult \acp{MCQ} subset (83.60\% pass@1, 86.40\% cons@16).

The other reasoning models also exhibit robust results, with performance remaining consistently above 85\% cons@16 overall. Notably, even the smaller reasoning model, Qwen3-4B, surpasses most larger non-reasoning models, emphasizing that reasoning-optimized architectures provide substantial gains in structured data understanding, independent of model size.

A key advantage of reasoning models is their resilience to complex questions, which often involve multi-step computations such as aggregations, averages, and conditional selections across multiple table entries. While non-reasoning models tend to struggle with these tasks, reasoning models sustain high accuracy.
A qualitative comparison of model responses on a representative MCQ is shown in Table~\ref{tab:comparison} (Appendix~\ref{app:qualitative}).

\subsubsection{Table-Specialized Models}
Table-specialized LLMs are explicitly trained to improve tabular reasoning,
either through supervised fine-tuning on large collections of structured
data or through reinforcement learning.

Table specialization has a highly variable effect on TeleTables, depending on both training paradigm and base model. TableGPT2-7B scores \emph{below} its Qwen2.5-7B base (46.40\% vs.\ 49.40\% cons@16), indicating that supervised fine-tuning on business intelligence tables introduces inductive biases misaligned with 3GPP specification tables~\citep{su2024tablegpt2}. Table-R1-Zero-7B, sharing the same base but trained with reinforcement learning with verifiable reward (RLVR), instead gains $+23.6$pp (73.00\% cons@16), suggesting that reinforcement learning generalizes more effectively to out-of-domain technical content. Most strikingly, both TableGPT-R1 and STaR-8B, built on Qwen3-8B, score below their unmodified base (87.20\% and 88.20\% vs.\ 89.60\%), confirming that table specialization on non-telecom data provides no benefit when the base model is already a strong reasoner.

\rev{In contrast, domain-specific fine-tuning on 3GPP data improves with-table performance (pass@1: 48.08\%\,$\to$\,59.00\%; cons@16: 49.40\%\,$\to$\,65.00\%). However, a substantial gap remains relative to the best general-purpose reasoning model (65.00\% vs.\ 92.60\% cons@16), indicating that domain specialization improves technical table understanding but does not substitute for strong reasoning capacity.}

\rev{\subsection{Difficulty is Structural and Systematic}
\label{sec:stratified}
Each question is annotated along two dimensions: \textit{reasoning skill} (the type of operation required, from direct lookup to multi-criteria aggregation) and \textit{evidence scope} (the portion of the table that must be consulted, from a single cell to the full table). We additionally stratify performance by \textit{table structure}, considering table size and row organization.
Table~\ref{tab:stratified} reports pass@1 for Qwen3-32B stratified by reasoning skill, evidence scope, and table structure.
Performance degrades systematically along all three axes. Across reasoning skills, accuracy spans 32.2\,pp, from 94.7\% on direct lookups to 62.5\% on multi-criteria questions, while increasing evidence scope and structural complexity are likewise associated with lower accuracy. These trends show that TeleTables \acp{MCQ} probe more than direct information retrieval, with difficulty arising from multiple aspects of table interpretation, including reasoning depth, evidence integration, and structural complexity.
Further analysis of how accuracy scales with table complexity, as measured by token count, is provided in Appendix~\ref{app:complexity}.}
\begin{table}[t]
\centering
\small
\caption{\rev{Pass@1 for Qwen3-32B by reasoning skill, evidence scope, and table structure.}}
\label{tab:stratified}
\begin{tabular}{lr}
\toprule
\textbf{Category} & \textbf{pass@1} \\
\midrule
\multicolumn{2}{l}{\textit{Reasoning skill}} \\
\quad Direct lookup  & 94.7\% \\
\quad Counting       & 92.4\% \\
\quad Arithmetic     & 93.3\% \\
\quad Comparison     & 86.6\% \\
\quad Aggregation    & 81.7\% \\
\quad Multi-criteria & 62.5\% \\
\addlinespace
\multicolumn{2}{l}{\textit{Evidence scope}} \\
\quad Single cell     & 93.2\% \\
\quad Cross-reference & 89.3\% \\
\quad Full-table scan & 86.2\% \\
\addlinespace
\multicolumn{2}{l}{\textit{Table structure}} \\
\quad Small (${\leq}$5$\times$4) & 100.0\% \\
\quad Large ({>}20r\,/\,8c)      &  88.0\% \\
\quad Grouped rows               &  89.5\% \\
\bottomrule
\end{tabular}
\end{table}

%%%%%%%%%%%%%%%%%%%%%%%%%%%%%%%%%%%%%%%%%%%%%%%%%%%%%
\section{Discussion}
\label{sec:discussion}
\vspace{-0.3cm}
\rev{The two evaluation settings expose distinct performance bottlenecks.}

\rev{In the \textit{closed-book} setting, domain knowledge is the primary constraint: no general-purpose model exceeds 40.6\% cons@16, while a 7B domain fine-tuned model reaches 58.0\%, outperforming all tested open-weight models despite being up to 17$\times$ smaller. In the \textit{with-table} setting, however, the same domain fine-tuned model reaches only 65.0\%, well below the best general-purpose reasoning model at 92.6\%. Together, these results suggest that domain specialization can compensate for missing domain knowledge, but does not substitute for strong reasoning capacity.}

\rev{This distinction is reinforced by the stratified analysis in Table~\ref{tab:stratified}. When the table is provided, performance degrades systematically with reasoning depth (94.7\% for direct lookup $\to$ 62.5\% for multi-criteria reasoning), evidence scope (93.2\% for single-cell evidence $\to$ 86.2\% for full-table evidence), and structural complexity (100\% for small tables $\to$ 88\% for large tables). Thus, once the relevant information is available in context, the remaining challenge lies in interpreting, integrating, and reasoning over increasingly complex tabular evidence.}

\rev{Table specialization alone does not consistently improve performance. Models fine-tuned on non-telecom tables (TableGPT2-7B, TableGPT-R1, and STaR-8B) score at or below their unmodified base models, whereas Table-R1-Zero-7B, which shares the same 7B base but is trained with RLVR, gains 23.6\,pp (73.00\% cons@16). This contrast suggests that the training objective and data distribution may be at least as important as specialization itself for transferring tabular reasoning capabilities to out-of-domain technical content.}

\rev{The results also reveal a notable interaction between representation and model modality. Multimodal models outperform their text-only counterparts even when no image is provided (e.g., Qwen2.5-VL-72B at 87.40\% vs.\ Qwen2.5-72B at 79.40\%), suggesting that vision-language pretraining may transfer beneficial representations to text-based table understanding. Yet rendered images remain the weakest input format: Qwen2.5-VL-72B drops from 85.6\% with HTML to 77.2\% with images. This indicates that, for complex technical tables, explicit markup provides structural cues that current models exploit more effectively than visual layout alone.}

%%%%%%%%%%%%%%%%%%%%%%%%%%%%%%%%%%%%%%%%%%%%%%%%%%%%%
\section{Conclusions}
\label{sec:conclusions}

\rev{This paper introduced TeleTables, a benchmark for assessing LLMs' ability to interpret and reason over tabular data in telecommunication standards. Our experiments reveal two distinct bottlenecks. In the closed-book setting, domain knowledge is the primary constraint: no general-purpose model exceeds 40.6\% cons@16, while a 7B domain fine-tuned model reaches 58.0\%. When the table is provided as context, interpretation becomes the primary constraint, with accuracy degrading systematically with reasoning depth, evidence scope, and structural complexity. \rev{HTML emerges as the strongest single representation; reasoning models consistently outperform table-specialized models trained on non-telecom data; and multimodal models show gains in text-based table understanding even without image input.} Overall, our results show that domain specialization can substantially reduce the knowledge gap, but strong reasoning capabilities remain essential for reliable interpretation of complex technical tables.}

%%%%%%%%%%%%%%%%%%%%%%%%%%%%%%%%%%%%%%%%%%%%%%%%%%%%%
\section{Limitations}
\label{sec:limitations}

TeleTables covers 13 3GPP specifications from Releases 18 and 19; coverage of other releases, standardization bodies (e.g., IETF and IEEE 802), and technical domains remains to be explored.
All source documents and MCQs are in English, so cross-lingual transfer to technical tables in other languages is out of scope.
At 500 MCQs, the benchmark provides a focused evaluation but may not capture the full diversity of 3GPP tables; future versions could increase coverage by sampling from a broader set of specifications.
Our zero-context experiment uses a closed-book setting; the interaction between pretraining gaps and retrieval quality in end-to-end RAG pipelines warrants further study.
\rev{A potential concern is that using models from the Qwen family in the question-generation process could introduce an evaluation bias in favor of Qwen models. However, the results do not indicate a systematic advantage for Qwen models: the strongest non-Qwen models, GPT-OSS-120B (91.80\% cons@16) and SeedOSS-36B (91.60\%), perform within 1\,pp of Qwen3-32B (92.60\%), while Llama3.3-70B (82.40\%) outperforms Qwen2.5-72B (79.40\%) at comparable scale. Moreover, all questions are strictly grounded in the source tables and were reviewed by domain experts before inclusion.}

%%%%%%%%%%%%%%%%%%%%%%%%%%%%%%%%%%%%%%%%%%%%%%%%%%%%%
\bibliography{reference}

@article{llm:telecom_impact,
  author={Maatouk, Ali and Ayed, Fadhel and Piovesan, Nicola and Domenico, Antonio De and Debbah, Merouane},
  journal={arXiv preprint arXiv:2308.06013}, 
  title={{Large Language Models for Telecom: Forthcoming Impact on the Industry}}, 
  year={2023}
}

@article{teleqna:paper,
  author={Maatouk, Ali and Ayed, Fadhel and Piovesan, Nicola and Domenico, Antonio De and Debbah, Merouane and Luo, Zhi-Quan},
  journal={IEEE Network}, 
  title={{TeleQnA: A Benchmark Dataset to Assess Large Language Models Telecommunications Knowledge}}, 
  year={2025},
  volume={},
  number={},
  pages={1-1},
  doi={10.1109/MNET.2025.3576035}
}

@article{nikbakht2024tspec,
  title={{TSpec-LLM: An Open-source Dataset for LLM Understanding of 3GPP Specifications}},
  author={Nikbakht, Rasoul and Benzaghta, Mohamed and Geraci, Giovanni},
  journal={arXiv preprint arXiv:2406.01768},
  year={2024}
}

@article{colle2024telemath,
  title={{TeleMath: A Benchmark for Large Language Models in Telecom Mathematical Problem Solving}},
  author={Colle, Vincenzo and Sana, Mohamed and Piovesan, Nicola and De Domenico, Antonio and Ayed, Fadhel and Debbah, Merouane},
  journal={arXiv preprint arXiv:2506.10674},
  year={2024}
}

@misc{roychowdhury2024evaluation,
      title={Evaluation of Table Representations to Answer Questions from Tables in Documents : A Case Study using 3GPP Specifications}, 
      author={Sujoy Roychowdhury and Sumit Soman and HG Ranjani and Avantika Sharma and Neeraj Gunda and Sai Krishna Bala},
      year={2024},
      eprint={2408.17008},
      archivePrefix={arXiv},
      primaryClass={cs.IR},
      url={https://arxiv.org/abs/2408.17008}, 
}

@article{sana2025reasoning,
  title={{Reasoning Language Models for Root Cause Analysis in 5G Wireless Networks}},
  author={Sana, Mohamed and Piovesan, Nicola and De Domenico, Antonio and Kang, Yibin and Zhang, Haozhe and Debbah, Merouane and Ayed, Fadhel},
  journal={arXiv preprint arXiv:2507.21974},
  year={2025}
}

@misc{zheng2024multimodal,
      title={Multimodal Table Understanding}, 
      author={Mingyu Zheng and Xinwei Feng and Qingyi Si and Qiaoqiao She and Zheng Lin and Wenbin Jiang and Weiping Wang},
      year={2024},
      eprint={2406.08100},
      archivePrefix={arXiv},
      primaryClass={cs.CL},
      url={https://arxiv.org/abs/2406.08100}, 
}

@inproceedings{wei2022chain,
 author = {Wei, Jason and Wang, Xuezhi and Schuurmans, Dale and Bosma, Maarten and Xia, Fei and Chi, Ed and Le, Quoc V and Zhou, Denny},
 booktitle = {Advances in Neural Information Processing Systems},
 pages = {24824--24837},
 title = {{Chain-of-Thought Prompting Elicits Reasoning in Large Language Models}},
 volume = {35},
 year = {2022}
}

@inproceedings{mctest2013,
  title={{MCTest: A Challenge Dataset for the Open-Domain Machine Comprehension of Text}},
  author={Richardson, Matthew and Burges, Christopher J.C. and Renshaw, Erin},
  booktitle={Proceedings of the 2013 Conference on Empirical Methods in Natural Language Processing},
  pages={193--203},
  year={2013}
}

@misc{telcoorag2025,
      title={Telco-oRAG: Optimizing Retrieval-augmented Generation for Telecom Queries via Hybrid Retrieval and Neural Routing}, 
      author={Andrei-Laurentiu Bornea and Fadhel Ayed and Antonio De Domenico and Nicola Piovesan and Tareq Si Salem and Ali Maatouk},
      year={2025},
      eprint={2505.11856},
      archivePrefix={arXiv},
      primaryClass={cs.IR},
      url={https://arxiv.org/abs/2505.11856}, 
}

@misc{chat3gpp,
      title={Chat3GPP: An Open-Source Retrieval-Augmented Generation Framework for 3GPP Documents}, 
      author={Long Huang and Ming Zhao and Limin Xiao and Xiujun Zhang and Jungang Hu},
      year={2025},
      eprint={2501.13954},
      archivePrefix={arXiv},
      primaryClass={cs.CL},
      url={https://arxiv.org/abs/2501.13954}, 
}

@misc{TeleTables,
author = {{Huawei Technologies}},
title = {{TeleTables}},
url={https://huggingface.co/datasets/netop/TeleTables}
}

@inproceedings{sui2024table,
  title={Table meets llm: Can large language models understand structured table data? a benchmark and empirical study},
  author={Sui, Yuan and Zhou, Mengyu and Zhou, Mingjie and Han, Shi and Zhang, Dongmei},
  booktitle={Proceedings of the 17th ACM International Conference on Web Search and Data Mining},
  pages={645--654},
  year={2024}
}

@inproceedings{wu2025tablebench,
  title={Tablebench: A comprehensive and complex benchmark for table question answering},
  author={Wu, Xianjie and Yang, Jian and Chai, Linzheng and Zhang, Ge and Liu, Jiaheng and Du, Xeron and Liang, Di and Shu, Daixin and Cheng, Xianfu and Sun, Tianzhen and others},
  booktitle={Proceedings of the AAAI Conference on Artificial Intelligence},
  volume={39},
  number={24},
  pages={25497--25506},
  year={2025}
}

@article{zhu2025tableeval,
  title={TableEval: A Real-World Benchmark for Complex, Multilingual, and Multi-Structured Table Question Answering},
  author={Zhu, Junnan and Wang, Jingyi and Yu, Bohan and Wu, Xiaoyu and Li, Junbo and Wang, Lei and Xu, Nan},
  journal={arXiv preprint arXiv:2506.03949},
  year={2025}
}

@article{su2024tablegpt2,
  title={Tablegpt2: A large multimodal model with tabular data integration},
  author={Su, Aofeng and Wang, Aowen and Ye, Chao and Zhou, Chen and Zhang, Ga and Chen, Gang and Zhu, Guangcheng and Wang, Haobo and Xu, Haokai and Chen, Hao and others},
  journal={arXiv preprint arXiv:2411.02059},
  year={2024}
}

@misc{yang2025tabler1,
      title={Table-R1: Inference-Time Scaling for Table Reasoning}, 
      author={Zheyuan Yang and Lyuhao Chen and Arman Cohan and Yilun Zhao},
      year={2025},
      eprint={2505.23621},
      archivePrefix={arXiv},
      primaryClass={cs.CL},
      url={https://arxiv.org/abs/2505.23621}, 
}

%%%%%%%%%%%%%%%%%%%%%%%%%%%%%%%%%%%%%%%%%%%%%%%%%%%%%
\onecolumn
\appendix

\section{Examples of Tables from 3GPP Standard Specifications}
\label{app:tabex}
We show a few examples of tables from 3GPP standard specifications, illustrating the structural variety and complexity that such tables can exhibit.

\begin{figure}[ht!]
    \centering
    \includegraphics[width=0.8\linewidth]{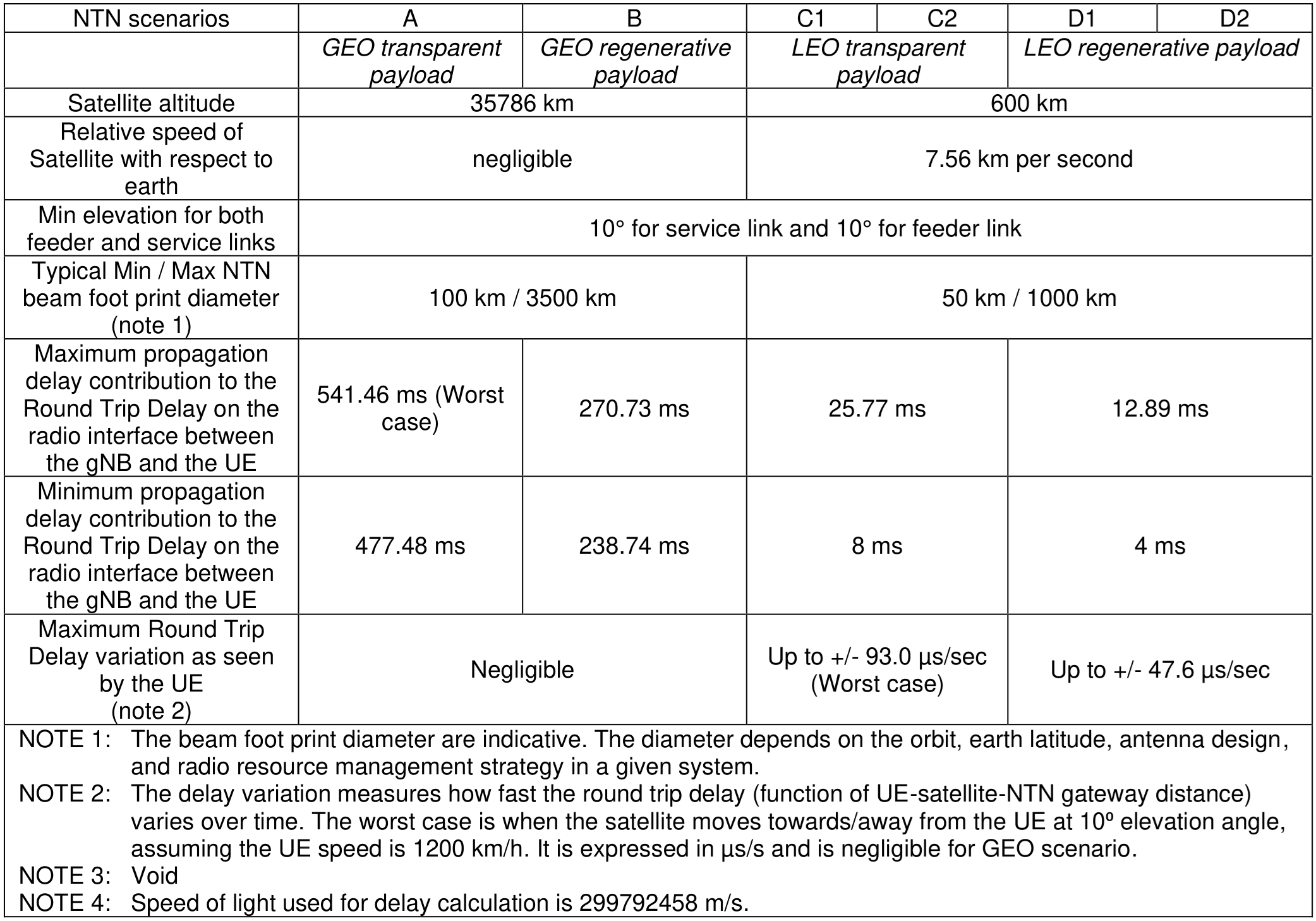}
    \caption{Table 7.1-1: NTN scenarios versus delay constraints from 3GPP TR 38.821 ("Solutions for NTN" / NR to support Non-Terrestrial Networks)}
    \label{fig:tabex1}
\end{figure}

Figure \ref{fig:tabex1} shows a table from 3GPP TR 38.821 ("Solutions for NTN" / NR to support Non-Terrestrial Networks) describing reference parameters for Non-Terrestrial Network (NTN) deployment scenarios. This example illustrates irregular header hierarchies and cell merging: the second header row groups scenario columns unevenly (C1/C2 share one label, D1/D2 share another, while A and B each have their own), and several body rows merge cells across columns in the opposite direction, with a single value spanning multiple scenarios (e.g. "35786 km" spanning columns A–B, "600 km" spanning C1–D2). The table also includes a substantial footnote block with numbered notes containing detailed prose explanations tied to specific rows.

\begin{figure}[ht!]
    \centering
    \includegraphics[width=0.8\linewidth]{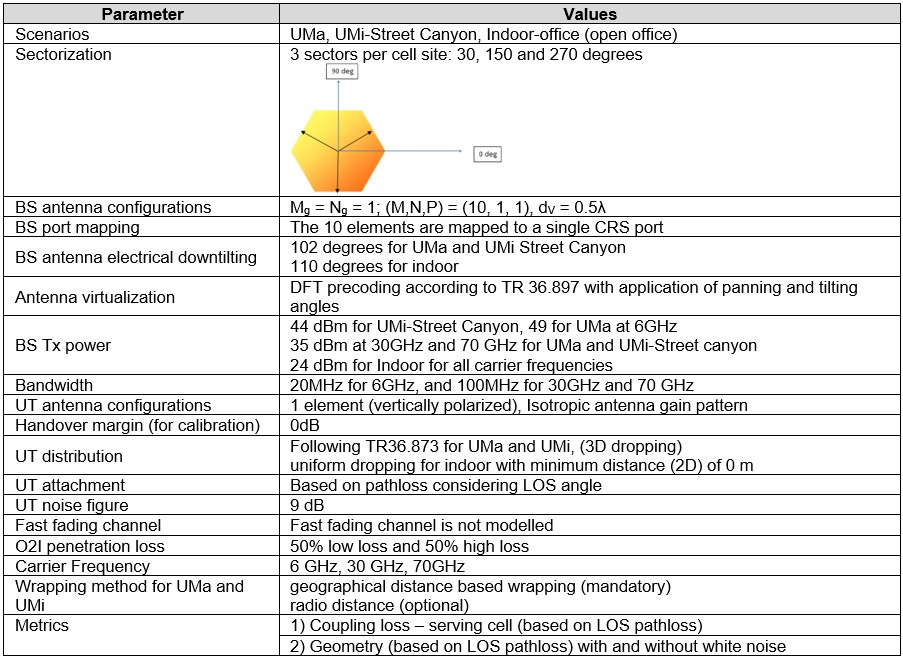}
    \caption{Table 7.8-1: Simulation assumptions for large scale calibration from 3GPP TR 38.901 (Study on channel model for frequencies from 0.5 to 100 GHz)}
    \label{fig:tabex2}
\end{figure}

Figure \ref{fig:tabex2} shows a table of simulation assumptions adapted from 3GPP TR 38.901 (Study on channel model for frequencies from 0.5 to 100 GHz), with a simple two-column "Parameter / Values" layout. Despite this simple structure, individual cells contain highly heterogeneous content: one cell embeds a raster image (a hexagonal sector diagram), others contain multi-line lists of conditional values, and one cell nests an enumerated sub-list. This illustrates how table complexity can reside entirely within cell content rather than in the overall grid structure.

\begin{figure}[ht!]
    \centering
    \includegraphics[width=0.8\linewidth]{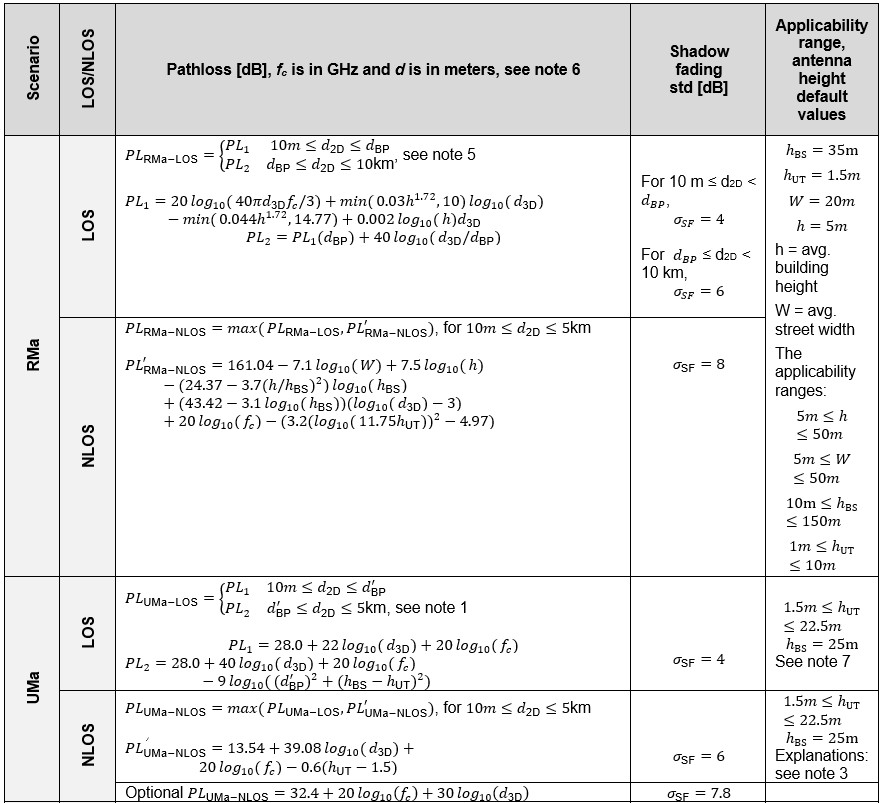}
    \caption{Part of Table 7.4.1-1: Pathloss models from 3GPP TR 38.901 (Study on channel model for frequencies from 0.5 to 100 GHz)}
    \label{fig:tabex3}
\end{figure}

Figure \ref{fig:tabex3} shows a pathloss model table from 3GPP TR 38.901 (Study on channel model for frequencies from 0.5 to 100 GHz), exhibiting nested row-spanning across two hierarchical levels (Scenario, then LOS/NLOS), alongside cells containing complete piecewise mathematical formulas with conditional branching, subscripts, and embedded variable glossaries. An additional row breaks the established row-span pattern, showing how exceptions to a table's own conventions are common in practice.

\newpage
\section{Dataset Examples}
\label{app:examples}

We show one example each of a basic and a difficult \ac{MCQ} produced by our generation pipeline, illustrating the range of reasoning difficulty in TeleTables.

\begin{tcolorbox}[title=\textbf{Basic MCQ},
  colframe=black, colback=white, coltitle=black,
  colbacktitle=gray!30, fonttitle=\bfseries,
  boxrule=0.2mm, width=\linewidth]
\textbf{Question:} What is the subcarrier spacing for reference channel G-FR2-A3-11?
\begin{enumerate}
    \item 90 kHz
    \item 180 kHz
    \item 30 kHz
    \item 120 kHz
    \item \textbf{60 kHz}
\end{enumerate}
\textbf{Explanation:} \textit{According to the table, the subcarrier spacing for G-FR2-A3-11 is 60 kHz.}

\textbf{Table:} \textit{Table A.3-9: FRC parameters for FR2 PUSCH performance requirements (\ldots)}

\textbf{Document:} \textit{3GPP TS 38.104 V19.0.0 (\ldots)}
\end{tcolorbox}

\vspace{6pt}

\begin{tcolorbox}[title=\textbf{Difficult MCQ},
  colframe=black, colback=white, coltitle=black,
  colbacktitle=gray!30, fonttitle=\bfseries,
  boxrule=0.2mm, width=\linewidth]
\textbf{Question:} Which reference channel experiences the smallest percentage reduction in total bits per slot when PT-RS is added?
\begin{enumerate}
    \item G-FR2-A7-1
    \item G-FR2-A7-2
    \item G-FR2-A7-3
    \item G-FR2-A7-4
    \item \textbf{All channels experience the same reduction}
\end{enumerate}
\textbf{Explanation:} \textit{The percentage reduction in bits when PT-RS is added is consistent across all channels. For example, G-FR2-A7-1 reduces from 57024 to 54648 bits, a 4.166\% decrease. The same ratio is observed in all channels, indicating a uniform impact of PT-RS configuration across the dataset.}

\textbf{Table:} \textit{Table A.7-1: FRC parameters for FR2-1 PUSCH performance requirements (\ldots)}

\textbf{Document:} \textit{3GPP TS 38.104 V19.0.0 (\ldots)}
\end{tcolorbox}

%%%%%%%%%%%%%%%%%%%%%%%%%%%%%%%%%%%%%%%%%%%%%%%%%%%%%
\newpage
\rev{\section{Accepted and Rejected Question Examples}
\label{app:rejected}
The following examples illustrate the acceptance and rejection criteria applied during human review.
An MCQ was accepted only if it was factually correct, representative of a question a domain expert would plausibly ask, and had exactly one defensible correct answer.}

\begin{tcolorbox}[title=\textbf{Rejected — Ambiguous answer (multiple correct options)},
  colframe=black, colback=white, coltitle=black,
  colbacktitle=gray!30, fonttitle=\bfseries,
  boxrule=0.2mm, width=\linewidth]
\textbf{Source:} Table A.2.1.1-1: E-UTRA simulation case minimum set (3GPP TS 36.814)

\textbf{Question:} Which simulation case has the highest total FDD bandwidth (sum of FDD uplink and downlink) when considering extended scenarios?

\textbf{Options:} (A) 3GPP case 1 extended, (B) ITU Indoor extended, (C) ITU Base coverage urban extended, (D) ITU Microcellular, (E) 3GPP case 3

\textbf{Rejection reason:} Options A, B, and C all yield 120\,MHz (80+40 MHz), making the answer non-unique.
\end{tcolorbox}

\vspace{6pt}

\begin{tcolorbox}[title=\textbf{Rejected — Correct answer absent from options},
  colframe=black, colback=white, coltitle=black,
  colbacktitle=gray!30, fonttitle=\bfseries,
  boxrule=0.2mm, width=\linewidth]
\textbf{Source:} Table 6.2.3.29-1: A-MPR regions for NS\_07 (3GPP TS 38.101-1)

\textbf{Question:} If a system operates at 10\,MHz bandwidth with $F_c = 782$\,MHz and $RB_\text{start} \times 12 \times \text{SCS} = 2.5$\,MHz, what is the minimum increase in $L_\text{CRB} \times 12 \times \text{SCS}$ required to move from A-MPR region A2 to A3?

\textbf{Options:} (A) 0\,MHz, \; (B) 1.44\,MHz, \; (C) 2.34\,MHz, \; (D) 3.24\,MHz, \; (E) 6.48\,MHz

\textbf{Rejection reason:} The correct value is $3.24 - 1.44 = 1.80$\,MHz, which does not correspond to any of the five options. Since no option is defensible, the question was rejected.
\end{tcolorbox}

\vspace{6pt}

\begin{tcolorbox}[title=\textbf{Accepted — Unambiguous, expert-plausible, table-grounded},
  colframe=black, colback=white, coltitle=black,
  colbacktitle=gray!30, fonttitle=\bfseries,
  boxrule=0.2mm, width=\linewidth]
\textbf{Source:} Table A.7-1: FRC parameters for FR2-1 PUSCH performance requirements (3GPP TS 38.104)

\textbf{Question:} Which reference channel experiences the smallest percentage reduction in total bits per slot when PT-RS is added?

\textbf{Correct answer:} All channels experience the same reduction.

\textbf{Acceptance reason:} Factually correct (uniform PT-RS overhead across all channels is verifiable from the table), representative of a question a standards engineer would ask when comparing channel configurations, and unambiguous.
\end{tcolorbox}

%%%%%%%%%%%%%%%%%%%%%%%%%%%%%%%%%%%%%%%%%%%%%%%%%%%%%
\newpage
\rev{\section{MCQ Generation Prompts}
\label{app:prompts}
We report the verbatim prompt templates used in each stage of the generation pipeline.}

\rev{\subsection*{Stage 1 --- Basic MCQ Generator}}

\begin{tcolorbox}[colframe=black, colback=gray!5, boxrule=0.2mm, width=\linewidth, fontupper=\small\ttfamily, breakable]
Analyze the table and create \{num\_questions\} multiple choice questions (MCQs).

For each MCQ, include: 1.\ Question text; 2.\ Five answer options (A--E); 3.\ Clearly marked correct answer key; 4.\ A detailed explanation justifying the correct answer.

The MCQs should: require understanding the table's structure and content; test extraction of specific information or simple trends; potentially require cross-referencing multiple data points; have exactly one clear, unambiguous, directly derivable correct answer; avoid options referring to other options by letter key.

Return a JSON array of \{num\_questions\} MCQs, each with fields: ``question'', ``options'' (dict A--E), ``answer\_key'', ``explanation''.
\end{tcolorbox}

\rev{\subsection*{Stage 1 --- Validator}}

\begin{tcolorbox}[colframe=black, colback=gray!5, boxrule=0.2mm, width=\linewidth, fontupper=\small\ttfamily, breakable]
You are given a table and a multiple-choice question about it. Your task is to answer the question based ONLY on the information in the table.

Question: \{question\_text\}\\
Options: \{options\_text\}

Please select the correct answer and provide a brief explanation. Respond as a JSON object with fields: ``answer'' (the option letter only) and ``explanation''.
\end{tcolorbox}

\rev{\subsection*{Stage 2 --- Difficult MCQ Generator}}

\begin{tcolorbox}[colframe=black, colback=gray!5, boxrule=0.2mm, width=\linewidth, fontupper=\small\ttfamily, breakable]
Role: You are an expert in telecommunications and a master question designer. Your task is to create exceptionally challenging MCQs for a benchmark designed to test the world's most advanced AI models.

CONTEXT: You have been provided with: (1) a data table in HTML and JSON formats; (2) a set of simple questions that test basic information retrieval from this table.

OBJECTIVE: Use the simple questions as inspiration to create \{questions\_per\_table\} much more difficult questions that require advanced reasoning. These questions should: synthesize information across multiple parts of the table; involve comparisons, trends, or patterns not explicitly stated; require mathematical operations; have exactly five options with one correct answer; include a detailed explanation.

Return a JSON array with fields: ``question'', ``options'' (A--E), ``answer\_key'', ``explanation''.
\end{tcolorbox}

\rev{\subsection*{Stage 2 --- Validator and Simplicity Filter}}

\rev{The Stage~2 validator (Qwen2.5-72B-Instruct) and simplicity filter (Qwen2.5-7B-Instruct) use the same prompt template.}

\begin{tcolorbox}[colframe=black, colback=gray!5, boxrule=0.2mm, width=\linewidth, fontupper=\small\ttfamily, breakable]
You are given a data table and a set of simple questions that were used as context to generate the following question. Your task is to carefully analyze all the provided information and answer the question precisely.

[DATA TABLE - HTML] \{table\_html\_data\} 

[DATA TABLE - JSON] \{table\_json\_data\}

[SIMPLE QUESTIONS CONTEXT] \{simple\_questions\_json\}

Answer the following multiple-choice question:\\
QUESTION: \{hard\_question\_text\}\\
OPTIONS: \{options\_text\}\\

Reason carefully, then put your selected option letter in $\backslash$boxed\{\} format. Example: $\backslash$boxed\{A\}
\end{tcolorbox}

%%%%%%%%%%%%%%%%%%%%%%%%%%%%%%%%%%%%%%%%%%%%%%%%%%%%%
\newpage
\section{Impact of Table Representation Formats}
\label{app:formats}

Table \ref{tab:input_formats} summarizes model performance across different table-representation formats and reports the average token usage associated with each format.
HTML yields the strongest performance for most models, particularly reasoning-capable models (GPT-OSS-120B: 91.4\%, Qwen3-32B: 92.0\%) and multimodal models (Qwen2.5-VL-72B: 85.6\%, Gemma-27B: 81.0\%).
We attribute this to the superior flexibility of HTML in representing complex table structures through explicit markup of hierarchical relationships and cell attributes.
The non-reasoning Qwen2.5-32B model also follows this trend, reaching its best performance of 85.6\% with HTML, while JSON and Markdown yield slightly lower results (83.6\% each). This supports the hypothesis that these models have a strong inductive bias toward HTML syntax from web-based pretraining data.
In contrast, the Llama models perform best with Markdown (Llama3.1-8B: 57.4\%, Llama3.3-70B: 80.2\%), likely reflecting their pretraining distribution. Notably, JSON achieves nearly identical performance to HTML for Llama3.3-70B (80.0\% vs.\ 79.8\%) while using 27\% fewer tokens (893 vs.\ 1224), indicating efficiency benefits without significant accuracy loss.

Multi-format concatenation generally improves performance, with the most pronounced gain for Llama3.1-8B (JSON+MD: +5.2pp). For larger models, improvements are modest (+0.6pp for Qwen3-32B), implying that capable architectures already capture structural relationships from a single format. The triple-format combination (HTML+JSON+MD) does not substantially outperform the best dual-format approach despite the increased token cost (2824 tokens).

Image representations perform substantially worse across all visual models: Qwen2.5-VL-72B drops from 85.6\% to 77.2\%, and Gemma-27B-it to 53.8\%. Hybrid inputs (e.g., IMAGE+HTML) recover most of this loss but at extreme token cost (12,831 tokens). Markdown uses 42\% fewer tokens than HTML while maintaining competitive performance, making it the preferred format for resource-constrained scenarios.

\begin{table*}[h]
\centering
\caption{Performance (\textit{cons@16}) of different LLMs across input formats.}
\label{tab:input_formats}
\resizebox{0.9\textwidth}{!}{%
\begin{tabular}{lc cccc ccc}
\toprule
\multirow{2}{*}{\textbf{Format}} & \multirow{2}{*}{\textbf{\shortstack{Avg.\ Tokens\\per Table}}} &
\multicolumn{3}{c}{\textbf{Non-reasoning}} &
\multicolumn{2}{c}{\textbf{Multimodal}} &
\multicolumn{2}{c}{\textbf{Reasoning}} \\
\cmidrule(lr){3-5} \cmidrule(lr){6-7} \cmidrule(lr){8-9}
& & \shortstack{Llama3.1\\8B} & \shortstack{Qwen2.5\\32B} & \shortstack{Llama3.3\\70B}
  & \shortstack{Gemma\\27b} & \shortstack{Qwen2.5-VL\\72B}
  & \shortstack{GPT-OSS\\120B} & \shortstack{Qwen3\\32B} \\
\midrule
HTML   & 1224  & 56.8 & \textbf{85.6} & 79.8 & 81.0 & 85.6  & 91.4 & 92.0 \\
JSON   &  893  & 56.8 & 83.6 & 80.0 & 77.6 & 85.4  & 91.0 & 90.4 \\
MD     &  707  & 57.4 & 83.6 & 80.2 & 79.6 & 84.0  & 90.2 & 90.4 \\
IMAGE  & 11607 & ---  & ---  & ---  & 53.8 & 77.2  & ---  & ---  \\
\midrule
HTML + JSON & 2117 & 61.0 & --- & \textbf{82.4} & 81.8 & \textbf{87.4} & 91.0 & \textbf{92.6} \\
HTML + MD   & 1931 & 60.2 & --- & 81.4 & \textbf{82.2} & 86.8 & 91.2 & 92.4 \\
JSON + MD   & 1600 & \textbf{62.6} & --- & 80.8 & 80.4 & 85.8 & 90.4 & 91.2 \\
IMAGE + HTML & 12831 & --- & --- & --- & 82.0 & 85.2 & --- & --- \\
\midrule
HTML + JSON + MD         & 2824  & 60.2 & --- & 81.4 & 82.2 & 86.6 & \textbf{91.8} & 92.6 \\
IMAGE + HTML + JSON + MD & 14431 & ---  & --- & ---  & 78.6 & 86.4 & ---  & ---  \\
\bottomrule
\end{tabular}
}
\end{table*}

\newpage
%%%%%%%%%%%%%%%%%%%%%%%%%%%%%%%%%%%%%%%%%%%%%%%%%%%%%
\section{Impact of Table Complexity}
\label{app:complexity}

We use the number of HTML tokens as a proxy for structural complexity (number of rows, columns, and merged cells). As shown in Figure~\ref{fig:performancePerTableTokens}, performance decreases consistently as complexity increases across all model types. Qwen3-32B achieves 100\% pass@1 on simple tables (fewer than 500 tokens) but drops to $\sim$80\% on the most complex ones. Qwen2.5-VL-72B-Instruct follows a similar trend with faster degradation, likely due to limited reasoning capability. Qwen2.5-7B-Instruct fails to reach satisfactory performance even on simple tables ($<$80\% pass@1), degrading to 27\% on the most complex ones.

\begin{figure}[h]
    \centering
    \begin{tikzpicture}
    \begin{axis}[
        xlabel={Table HTML Tokens},
        ylabel={Avg.\ pass@1 [\%]},
        width=0.92\columnwidth, height=7cm,
        grid=both,
        legend style={
            at={(0.98,0.98)},
            anchor=north east,
            draw=none,
            legend columns=1,
            font=\footnotesize,
        },
        legend cell align={left},
        xmax=2500]
    \addplot[mark=*] coordinates {
        (204.8995, 79.7500)
        (414.4000, 61.7500)
        (587.1000, 53.7500)
        (738.1000, 43.0147)
        (876.7500, 50.6378)
        (1034.9500, 42.6250)
        (1235.7000, 45.0721)
        (1527.5000, 40.4337)
        (2163.1000, 26.6582)
    };
    \addlegendentry{Qwen2.5-7B}
    \addplot[mark=*,color=red] coordinates {
        (204.8995, 100)
        (414.4000, 100)
        (587.1000, 95.5)
        (738.1000, 80.5147)
        (876.7500, 84.1837)
        (1034.9500, 78.3750)
        (1235.7000, 78.6058)
        (1527.5000, 57.3980)
        (2163.1000, 76.1480)
    };
    \addlegendentry{Qwen2.5-VL-72B}
    \addplot[mark=*,color=blue] coordinates {
        (204.8995, 100)
        (414.4000, 100)
        (587.1000, 98.1250)
        (738.1000, 88.2353)
        (876.7500, 95.1531)
        (1034.9500, 90.1250)
        (1235.7000, 84.3750)
        (1527.5000, 78.9541)
        (2163.1000, 86.8622)
    };
    \addlegendentry{Qwen3-32B}
    \end{axis}
    \end{tikzpicture}
    \caption{Average pass@1 performance across MCQs grouped by table complexity (HTML token count).}
    \label{fig:performancePerTableTokens}
\end{figure}

%%%%%%%%%%%%%%%%%%%%%%%%%%%%%%%%%%%%%%%%%%%%%%%%%%%%%
\section{Qualitative Model Comparison}
\label{app:qualitative}

Table~\ref{tab:comparison} shows the responses produced by four representative models on the same TeleTables \ac{MCQ}, illustrating how reasoning capability and model size affect the quality of tabular interpretation.

\begin{table*}[p]
\centering
\small
\caption{Responses from models of different types on the same TeleTables MCQ. The correct answer is Option~3 (index 513).}
\label{tab:comparison}
\begin{tabular}{p{2cm} p{10.5cm} p{2cm}}
\toprule
\textbf{Question} & Which of the following bit field indices indicate a configuration with 4 layers when the transform precoder is disabled and maxRank = 4? & \\
\textbf{Options} & Option 1: 117, \; Option 2: 1022, \; Option 3: 513, \; Option 4: 10, \; Option 5: 423 & \\
\textbf{Table} & Table 7.3.1.1.2-5L from 3GPP TS 38.212 \\ &\includegraphics[scale=0.13]{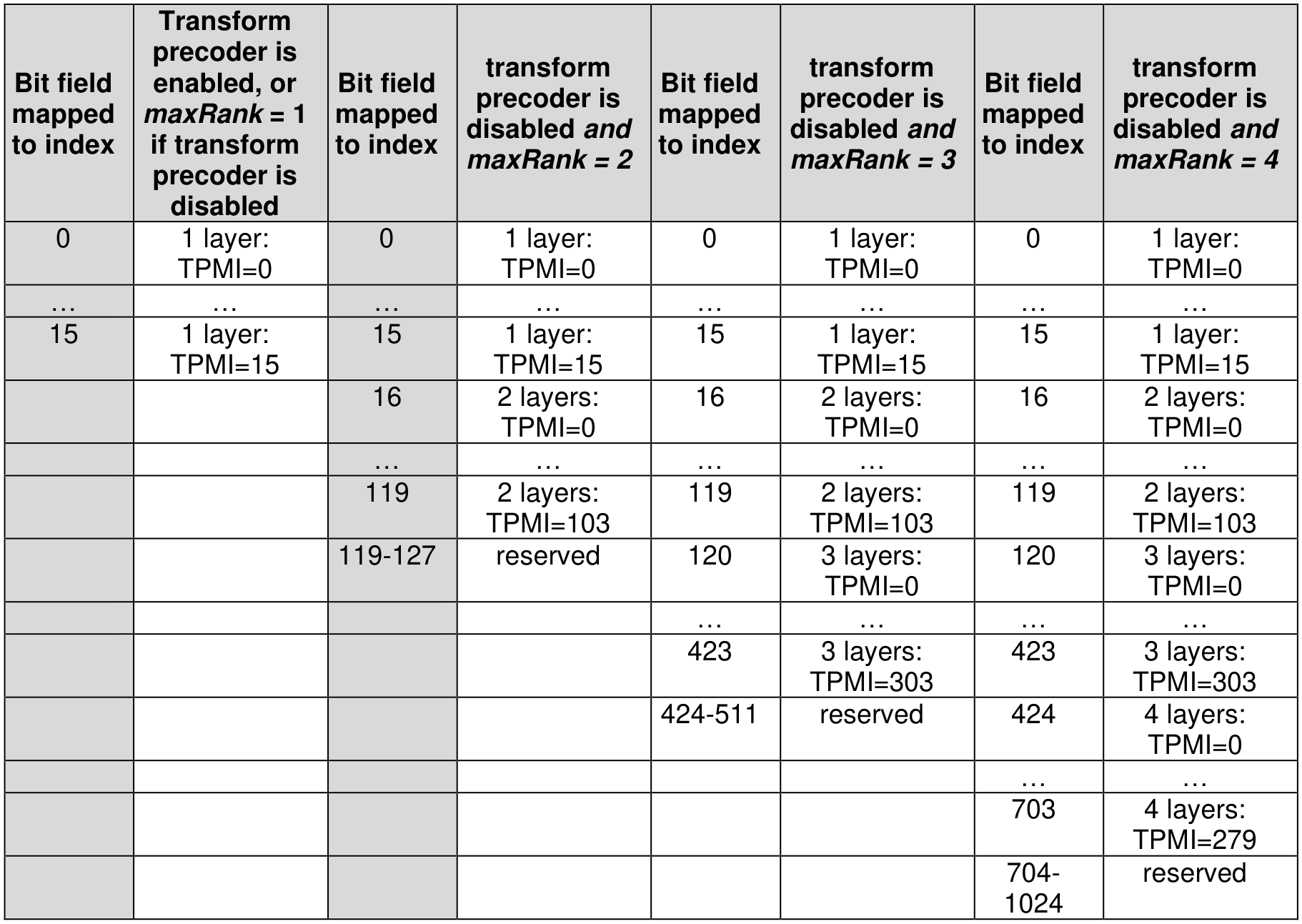} \\
\midrule
\textbf{Model} & \textbf{Answer Provided} & \textbf{Result} \\
\midrule
Qwen2.5-7B-Instruct & Based on the information provided in the table, the correct index for 4 layers when the transform precoder is disabled and maxRank = 4 should be around 704-1024, but none of the options match this range exactly. The closest provided option is Option 5: 423, which does not fit the criteria based on the table. $\boxed{704}$
 & \textcolor{red}{Wrong} \\
\addlinespace

Qwen2.5-VL-72B-Instruct &
To determine which bit field index indicates
a configuration with 4 layers when the transform precoder is disabled
and maxRank = 4, we need to refer to the relevant column in the provided
table. (\ldots)
From the table, we see the following ranges and values:
Bit field index 0 to 15 corresponds to 1 layer.
Bit field index 16 to 119 corresponds to 2 layers.
\textbf{Bit field index 424 to 703 corresponds to 4 layers.}
Bit field index 704 to 1024 is reserved. (\ldots)
Option 1: 117 - This falls within the range for 2 layers (16 to 119).
Option 2: 1022 - This falls within the reserved range (704 to 1024).
\textbf{Option 3: 513 - This does not fall within any specified range in the table for 4 layers.}
Option 4: 10 - This falls within the range for 1 layer (0 to 15).
Option 5: 423 - This is just before the range for 4 layers starts (424 to 703).
The only option that falls within the range for 4 layers is Option 2: 1022.
$\boxed{2}$
 & \textcolor{red}{Wrong} \\
\addlinespace

Qwen3-32B &
$\langle$think$\rangle$
(\ldots) The question specifies that the transform precoder is disabled
and maxRank is 4. \textbf{So I should look at the last column, which is
labeled transform precoder is disabled and maxRank = 4.} (\ldots)
\textbf{So for 4 layers, the valid indices are from 424 to 703.}
Now, looking at the options:
Option 1: 117 - This is below 424, so probably for 2 or 3 layers or reserved.
Option 2: 1022 - 1022 is above 703 and within the 704-1024 reserved range. So that's reserved.
\textbf{Option 3: 513 - 513 is between 424 and 703. So that's valid for 4 layers.}
Option 4: 10 - Way below 424, so no.
Option 5: 423 - This is the upper limit for 3 layers. The next index for 4 layers starts at 424. So 423 isn't in the 4-layer range.
Therefore, the correct option is Option 3: 513.
$\langle$/think$\rangle$
$\boxed{3}$
 & \textcolor{green!80!black!100}{Correct} \\
\addlinespace

STaR-8B &
$\langle$think$\rangle$
I need to find the row where transform precoder is disabled and maxRank = 4.
From the table:
\textbf{| ... | 423 | 3 layers: TPMI = 303 | 424 | 4 layers: TPMI = 0 | ... |}
So index 424 corresponds to 4 layers. However, 424 is not among the
options. (\ldots) The closest option is 423 (Option 5), which maps to
3 layers. (\ldots) Since none of the options exactly match index 424,
I will assume the question intended 3 layers and select 423.
$\langle$/think$\rangle$ $\boxed{4}$
& \textcolor{red}{Wrong} \\
\bottomrule
\end{tabular}
\end{table*}

%% Acronym definitions (not printed)
\begin{acronym}[AAAAAAAAA]
  \acro{3GPP}{Third Generation Partnership Project}
  \acro{AI}{Artificial Intelligence}
  \acro{LLM}{large language model}
  \acro{MCQ}{multiple choice question}
  \acro{ML}{Machine Learning}
  \acro{NLP}{Natural Language Processing}
  \acro{RAG}{Retrieval Augmented Generation}
\end{acronym}

\end{document}